\documentclass{article}

\usepackage{microtype}
\usepackage{graphicx}
\usepackage{subcaption}
\usepackage{booktabs} 

\usepackage{xcolor}
\usepackage{multirow}
\usepackage{listings}
\usepackage{amsmath,amssymb,amsthm,amsfonts}
\usepackage{eqnarray}
\usepackage{multicol}
\usepackage{graphicx}

\usepackage{hyperref}

\usepackage[accepted]{icml2020}

\icmltitlerunning{Learning Fair Rule Lists}

\begin{document}

\newtheorem{definition}[theorem]{Definition} 

\def\classifier{\mathcal{C}}

\def\fairl{\texttt{FairCORELS}}
\def\laftr{\texttt{LAFTR}}
\def\zafar{\texttt{C-LR}}
\def\faht{\texttt{FAHT}}

\def\corels{\texttt{\textsc{CORELS}}}
\def\fairml{\texttt{FairML}}
\def\ParetoFront{\texttt{ParetoFrontGeneration}}
\def\ULB{\texttt{ULB}}

\def\dataset{\mathcal{D}}
\def\nbdata{N}
\def\dimdata{P}
\def\legitimate{X}
\def\adataind#1{x_{#1}}
\def\adata{x}\def\adata{x}

\def\protected{G} 
\def\decision{Y}
\def\adecind#1{y_{#1}}
\def\adec{y}
\def\datatuple{(\legitimate{},\protected{},\decision{})}
\def\trainingset{(\legitimate{},\decision{})}
\def\rulelist{r} 
\def\lenrl{K}
\def\oneant#1{p_{#1}}
\def\oneconseq#1{q_{#1}}
\def\defconseq{q_0}
\def\rl{r=(\{\oneant{k} / k \in 1\ldots\lenrl \}, \{\oneconseq{k} / k \in 1 \ldots\lenrl \}, \defconseq)}

\def\arule{r}
\def\regul{\lambda}
\def\misc{\mathsf{Error}}
\def\acc{\mathsf{Accuracy}}
\def\lwb{\mathsf{b}}

\def\obj{\texttt{Obj}}

\def\setrules{S}

\def\unfair#1{\mathsf{UNF_{#1}}}
\def\incons#1{inconsistency^{#1}}
\def\statparity{SP}
\def\predparity{PP}
\def\predequality{PP}
\def\eqopportunity{EOpp}
\def\eqodds{EOdds}
\def\cond{CUAE}

\def\pred#1{pred_{#1}} 
\def\pos#1{pos_{#1}}   

\def\MObj{Z}
\def\nbobj{m}
\def\oneobj#1{z_{#1}}
\def\setvar{V}
\def\setcts{\Omega}
\def\onesol{v}

\def\coefWSum{\beta}
\def\coefEps{\epsilon}
\def\NbPtPareto{E}
\def\unfairness{\mathsf{unfairness}}

\def\msp{\mathsf{\unfair{\statparity}}}
\def\meopp{\mathsf{\unfair{\eqopportunity}}}
\def\meodds{\mathsf{\unfair{\eqodds}}}

\definecolor{codegreen}{rgb}{0,0.6,0}
\definecolor{codegray}{rgb}{0.5,0.5,0.5}
\definecolor{codepurple}{rgb}{0.58,0,0.82}
\definecolor{backcolour}{rgb}{0.95,0.95,0.92}
 
\lstdefinestyle{mystyle}{
    backgroundcolor=\color{backcolour},   
    commentstyle=\color{codegreen},
    keywordstyle=\bfseries\color{codepurple},
    numberstyle=\tiny\color{codegray},
    stringstyle=\color{codepurple},
    basicstyle=\ttfamily\tiny,
    breakatwhitespace=false,         
    breaklines=true, 
    captionpos=b,                    
    keepspaces=true,                 
    numbers=left,                    
    numbersep=5pt,                  
    showspaces=false,                
    showstringspaces=false,
    showtabs=false,                  
    tabsize=2
}
 
\lstset{style=mystyle, columns=fullflexible}
\renewcommand{\lstlistingname}{Rule list}

\twocolumn[
\icmltitle{Learning Fair Rule Lists}
\begin{icmlauthorlist}
\icmlauthor{Ulrich A{\"\i}vodji}{uqam}
\icmlauthor{Julien Ferry}{laas}
\icmlauthor{S{\'e}bastien Gambs}{uqam}
\icmlauthor{Marie-Jos{\'e} Huguet}{laas,tls}
\icmlauthor{Mohamed Siala}{laas,tls}
\end{icmlauthorlist}
\icmlaffiliation{uqam}{Universit{\'e} du Qu{\'e}bec {\`a} Montr{\'e}al}
\icmlaffiliation{laas}{LAAS-CNRS}
\icmlaffiliation{tls}{Universit{\'e} de Toulouse}
\icmlcorrespondingauthor{Ulrich A{\"\i}vodji}{aivodji.ulrich@courrier.uqam.ca}
\icmlcorrespondingauthor{Julien Ferry}{jferry@laas.fr}
\icmlcorrespondingauthor{S{\'e}bastien Gambs}{gambs.sebastien@uqam.ca}
\icmlcorrespondingauthor{Marie-Jos{\'e} Huguet}{huguet@laas.fr}
\icmlcorrespondingauthor{Mohamed Siala}{msiala@laas.fr}

\icmlkeywords{Machine Learning, Fairness, Interpretability}
\vskip 0.3in

]

\printAffiliationsAndNotice{}

\begin{abstract}
As the use of black-box models becomes ubiquitous in high stake decision-making systems, demands for fair and interpretable models are increasing. 
While it has been shown that interpretable models can be as accurate as black-box models in several critical domains, existing fair classification techniques that are interpretable by design often display poor accuracy/fairness tradeoffs in comparison with their non-interpretable counterparts. 
In this paper, we propose \fairl{}, a fair classification technique interpretable by design, whose objective is to learn fair rule lists. 
Our solution is a multi-objective variant of \corels{}, a branch-and-bound algorithm to learn rule lists, 
that supports several statistical notions of fairness. Examples of such measures include statistical parity, equal opportunity and equalized odds. 
The empirical evaluation of \fairl{} on real-world datasets demonstrates that it outperforms state-of-the-art fair classification techniques that are interpretable by design while being competitive with non-interpretable ones.
\end{abstract}

\section{Introduction}

Machine learning models are now becoming more and more common in high stake decision-making systems (\emph{e.g.}, credit scoring~\cite{siddiqi2012credit}, predictive justice~\cite{kleinberg2017human} and automatic recruiting~\cite{miller_2015}).
These systems can have an important impact on individuals as decisions based on wrong predictions can adversely affect human lives (\emph{e.g.}, people being wrongly denied parole~\cite{wexler2017computer}). 
Thus, ethical aspects such as the fairness and transparency of machine learning models have become not only desirable features but also legal requirements. 
For instance, the European General Data Protection Regulation (GDPR) has a provision requiring explanations of the rationale for decisions taken by automated systems (often based on machine learning models) that have a significant impact on individuals~\cite{goodman2016european}. 
In fact, understanding these models can be considered as a prerequisite towards quantitatively evaluating other criteria such as fairness, reliability and robustness ~\cite{doshi2017towards,bostrom2014ethics}.
Two main approaches have emerged in the literature to facilitate the understanding of machine learning models: \emph{black-box explanation} and \emph{transparent-box design}~\cite{lipton2016mythos,lepri2017fair,montavon2018methods,guidotti2018survey}. 

Black-box explanation techniques, also referred to as \emph{post-hoc explanations}, refer to methods designed to explain how black-box ML models produce their outcomes. Current existing techniques for post-hoc explanations include \emph{global explanations}~\cite{craven1996extracting,lakkaraju2017interpretable}, \emph{local explanations}~\cite{ribeiro2016should}, \emph{explanations by example}~\cite{mothilal2019explaining}, \emph{text explanations}~\cite{lei2016rationalizing}, \emph{visual explanations}~\cite{simonyan2013deep,selvaraju2016grad}, and \emph{feature relevance explanations}~\cite{vidovic2016feature}\footnote{We refer the interested reader to the following surveys~\cite{guidotti2018survey,arrieta2020explainable} for a detailed overview of these techniques.}.
However, while black-box explanations can be useful in debugging tasks or non-sensitive contexts, recent works~\cite{rudin2018please,aivodji2019fairwashing,fukuchi2019pretending,laugel2019dangers,heo2019fooling,dombrowski2019explanations,merrer2019bouncer,lakkaraju2019fool,slack2019can} suggest that they might be inappropriate in high-stake decision systems due to the fact that they can be arbitrarily manipulated to tell a different story than that of the black-box model they are explaining. 
For instance, local and global explanation techniques can be manipulated to under-report the unfairness of the black-box model~\cite{aivodji2019fairwashing}.

In contrast, transparent-box design aim at building transparent models, which are inherently interpretable~\cite{li2002mining,angelino2018learning,breiman2017classification,ustun2016supersparse}. 
For instance, when they have small or of reasonable size~\cite{lipton2016mythos}, the following models can be considered as interpretable: rules sets~\cite{rijnbeek2010finding,mccormick2011hierarchical,dash2018boolean,li2002mining}, rule lists~\cite{angelino2017learning,yang2017scalable,wang2015falling}, decision trees~\cite{breiman2017classification,narodytska2018learning} and scoring systems~\cite{zeng2017interpretable,ustun2016supersparse,koh2006two}. 
More recently, an hybrid framework for transparency~\cite{wang2019gaining,rafique2019model} has been proposed to associate black-box models with their interpretable partial substitutes. 

With respect to fairness, a significant amount of work has been done in recent years to design fairness-aware machine learning models~\cite{friedler2019comparative,DBLP:conf/fat/CelisHKV19}, which we will review in Section~\ref{sec:related}. 
Nonetheless, despite the progress made in both directions, fair classification techniques that are interpretable by design often display poor accuracy/fairness tradeoffs in comparison with their non-interpretable counterparts. 
To address this issue, we propose \fairl{}, a supervised learning algorithm whose objective is to build rule lists models that are both fair and accurate. 
\fairl{} leverages on recent advancements, provided by \corels{}, for learning certifiably optimal rule lists~\cite{angelino2017learning,angelino2018learning}, by adapting them to also take into account fairness constraints. 
In particular, given a statistical notion of fairness and a sensitive attribute that could lead to discrimination, our algorithm searches for the rule list optimizing the decrease of both unfairness and misclassification error. 

Our main contributions can be summarized as follows.
\begin{itemize}
\item We formulate the problem of learning fair rule lists as a multi-objective version of the problem addressed by \corels{}~\cite{ angelino2017learning,angelino2018learning}.
Afterwards, we propose \fairl{}, a supervised learning algorithm designed to build fair rule lists with high accuracy.
\item We evaluate \fairl{} on two public datasets, using six different statistical notions of fairness, namely statistical parity, predictive parity, predictive equality,  equal opportunity, equalized odds and conditional use accuracy equality. 
\item We compare its performances to both interpretable and non-interpretable state-of-the-art fair classification techniques.
This evaluation demonstrates that it outperforms existing fair and interpretable methods, while performing similarly to existing non-interpretable fair methods.
\end{itemize}

The outline of the paper is as follows.
First in Section~\ref{sec:preliminaries}, we review the background notions on fairness, rule lists, and multi-objective optimization necessary to the understanding of our work. 
Then, in Section~\ref{sec:related}, we present the related work on fairness-enhancing techniques. 
Afterwards in Section~\ref{sec:fairrulelists}, we introduce our multi-objective optimization framework for learning fair rule lists before describing \fairl{}, our learning algorithm for realizing this task. 
Finally, we report on the experiments conducted in Section~\ref{sec:experiments} before concluding in Section~\ref{sec:conclusion}.

\section{Preliminaries}
\label{sec:preliminaries}

In this section, we review the notions necessary to the understanding of our work, namely the fairness metrics, the rule lists and \corels{} as well as multi-objective optimization.

\subsection{Fairness Metrics}
\label{subsec:fairness}

Given a training dataset composed of features vectors $X \in \mathbb{R}^n$ (\emph{i.e.}, typically profiles of individuals), labels $Y \in \{0, 1\}$ (\emph{i.e.}, the class to predict) and sensitive attributes $A \in \{0, 1\}$ that could lead to discrimination, a fair learning algorithm for classification aims at producing a model whose predictions $\hat{Y}$ are simultaneously accurate with respect to $Y$ and ``fair'' with respect to demographic groups characterized by $A$. 
The literature on fairness in machine learning has boomed in recent years, thus it would be impossible to provide a complete review of the existing fairness notions in this paper. 
We refer the interested reader to the following surveys~\cite{narayanan2018translation,berk2017fairness,verma2018fairness,chouldechova2018frontiers} for a detailed overview of these notions. 

In a nutshell, three families of fairness notions have emerged: \emph{causal notions of fairness}~\cite{kilbertus2017avoiding,kusner2017counterfactual,nabi2018fair}, which rely on causal assumptions to estimate the effects of sensitive attributes and to build algorithms ensuring a low level of discrimination with respect to these attributes, \emph{individual notions of fairness}~\cite{dwork2012fairness,joseph2016fairness}, which lead to the same decisions for similar individuals, and \emph{statistical notions of fairness}~\cite{calders2010three,chouldechova2017fair,corbett2017algorithmic,hardt2016equality}, which require the model to exhibit approximate parity for some statistical measure across the different groups defined by $A$. 

We define the following six statistical measures of fairness:
statistical parity~\cite{dwork2012fairness,calders2010three,kamishima2011fairness,feldman2015certifying,zliobaite2015relation}, predictive parity~\cite{chouldechova2017fair,kleinberg2017inherent}, predictive equality~\cite{chouldechova2017fair,corbett2017algorithmic}, equal opportunity~\cite{hardt2016equality}, equalized odds~\cite{chouldechova2017fair,kleinberg2017inherent,hardt2016equality,zafar2017fairness} and conditional use accuracy equality~\cite{berk2017fairness}.

\begin{definition}[Statistical parity]
\label{def:sp}
\emph{Statistical parity, also known as demographic parity, requires the positive outcome to be given at the same rate for both groups. The statistical parity metric is defined as:}
{\scriptsize
\begin{align*}
    \unfair{SP} &= |P(\hat{Y}=1|A=0) - P(\hat{Y}=1|A=1)|.
\end{align*}
}%
\end{definition}

\begin{definition}[Predictive parity]
\label{def:pp}
\emph{Predictive parity requires the same positive predictive value (\emph{i.e.}, precision) in both groups. 
The predictive parity metric is defined as:}
{\scriptsize
\begin{align*}
    \unfair{PP} &= |P(Y=1|\hat{Y}=1,A=0) - P(Y=1|\hat{Y}=1,A=1)|.
\end{align*}
}%
\end{definition}

\begin{definition}[Predictive equality]
\label{def:pe}
\emph{Predictive equality requires the same false positive rate in both groups. 
The predictive equality metric is defined as:}
{\scriptsize
\begin{align*}
    \unfair{PE} &= |P(\hat{Y}=1|Y=0,A= 0) - P(\hat{Y}=1|Y=0,A= 1)|.
\end{align*}
}%
\end{definition}

\begin{definition}[Equal opportunity]
\label{def:eopp}
\emph{Equal opportunity requires the same true positive rate in both groups. 
The equal opportunity metric is defined as:}
{\scriptsize
\begin{align*}
    \unfair{EOpp} &= |P(\hat{Y}=1|Y=1,A= 0) - P(\hat{Y}=1|Y=1,A= 1)|.
\end{align*}
}%
\end{definition}

\begin{definition}[Equalized odds]
\label{def:eodds}
\emph{Equalized odds requires the same true positive rate and the same false positive rate in both groups. 
The equalized odds metric is defined as:} 
{\scriptsize
\begin{align*}
    \unfair{EOdds} &= |P(\hat{Y}=1|Y=1,A=0) - P(\hat{Y}=1|Y=1,A=1)| \\
    & + |P(\hat{Y}=1|Y=0,A=0) - P(\hat{Y}=1|Y=0,A=1)|
\end{align*}
}%
\end{definition}

\begin{definition}[Conditional use accuracy equality]
\label{def:eodds}
\emph{Conditional use accuracy equality requires the same positive predictive value and the same negative predictive value in both groups. 
The conditional use accuracy equality metric is defined as:}
{\scriptsize
\begin{align*}
    \unfair{CUAE} &= |P(Y=1|\hat{Y}=1,A=0) - P(Y=1|\hat{Y}=1,A=1)| \\
    & + |P(Y=0|\hat{Y}=0,A=0) - P(Y=0|\hat{Y}=0,A=1)|.
\end{align*}
}%
\end{definition}

\subsection{Rule Lists and \corels{}}
\label{subsec:rulelist}
\emph{Rule lists}~\cite{rivest1987learning,angelino2018learning} 
(also known as \emph{decision lists}) are classifiers formed by an ordered list of \emph{if-then} rules with antecedents in the \emph{if} clauses and predictions in the \emph{then} clauses. 
For instance, Rule list~\ref{rl:rl_example} has been learned on the Adult dataset~\footnote{http://mlr.cs.umass.edu/ml/datasets/Adult} to predict the salary category. 

\begin{minipage}{\linewidth}
\begin{lstlisting}[language=Haskell,numbers=none, caption={Example of a rule list 
found by \fairl{} on Adult dataset.}, label=rl:rl_example]
if [capitalGain:>=5119] then [high]
else if [maritalStatus:single] then [low]
else if [workclass:private &&  education:bachelors] then [high]
else if [education:masters_doctorate] then [high]
else if [capitalLoss:1882-1978] then [high]
else [low]
\end{lstlisting}
\end{minipage}
\noindent 
More precisely, a rule list $\rl$ 
consists of $\lenrl$ distinct association rules $\oneant{k} \to \oneconseq{k}$, in which $\oneant{k}$
is the antecedent of the association rule and $\oneconseq{k}$
its associated consequent, followed by a default prediction $\defconseq$. 
To classify a new data point, the rules are applied sequentially until one rule triggers, in which case the associated prediction is reported. 
If no rule is triggered, then the default prediction -- which typically predicts the majority class -- is reported.
As shown in \citep{rivest1987learning}, rule lists generalize decision trees. 
More precisely for a given size (\emph{i.e.}, the depth of a decision tree or the maximum width of a rule for a rule list), rules lists are strictly more expressive than decision trees. 
As a consequence we can obtain more compact models using rule lists, which leads to higher interpretability.

\corels{}~\cite{angelino2018learning} is a supervised learning algorithm~\cite{liu1998integrating} that, given a training set, outputs the rule list minimizing the training loss function. 
\corels{} represents the search space of rule lists as a trie (\emph{i.e.}, prefix tree) formed by pre-mined rules from the training set and uses branch-and-bound techniques to find the optimal rule list. 
For a given rule list $\arule$, the objective function, denoted by $\obj{}$, to minimize is:
\begin{equation}
		\obj{}(\arule,\legitimate{},\decision{})= \misc(\arule,\legitimate{},\decision{}) + 
		 \regul . \lenrl,
\label{eq:objCorels}
\end{equation}
in which $\misc(\arule,\legitimate{},\decision{})=(1-\acc(\arule,\legitimate{},\decision{}))$ is the classification error and $\regul \geq 0$ the regularization parameter used to penalize longer rule lists. 
The first part of the objective function aims at obtaining accurate rule list whereas the second one has for objective to reduce the size of the optimal rule list, thus limiting over-fitting while also improving interpretability.
\corels{} proposes various search strategies and leverages on a collection of bounds to efficiently prune the search space. 
In particular, \corels{} implements breadth-first search, depth-first search and best-first search. 
The best-first search uses a priority queue that can be ordered with three different priorities: lower bound, objective or curiosity. 
The curiosity priority is proportional to the ratio of the objective lower bound of the prefix tree to its normalized support.

\subsection{Multi-objective Optimization}
Optimizing a decision process is the task of choosing a particular solution among a set of alternatives.
More precisely, each alternative belongs to the set of feasible solutions and its quality can be assessed through a given objective function.
In mono-objective optimization, the aim is to find the global optimum, which corresponds to a solution having the best value for this objective function.
However, in many applications, one has to consider the concurrent optimization of several objective functions. 

Contrary to mono-objective optimization, solving a multi-objective optimization problem produces a set of solutions offering a trade-off between the different objective functions.
The comparison between solutions is usually done through a dominance relation (\emph{e.g.}, Pareto dominance Definition~\ref{def:pareto}). 
The set of non-dominated solutions of a multi-objective optimization problem forms a Pareto frontier.
\begin{definition}[Pareto dominance]
\label{def:pareto}
A solution $\onesol$ dominates a solution $\onesol'$ if $\onesol$ is at least as good as $\onesol'$ for all objectives and $\onesol$ is strictly better than $\onesol'$ for at least one objective.
\end{definition}
 
Various approaches exist for solving multi-objective optimization problems~\cite{ehrgott2005multicriteria, collette2013multiobjective}. 
In this paper, we focus on the $\epsilon$-constraint method, which aims at optimizing only one of the objective functions given a set of constraints on the others. 
The formulation of the multi-objective problem is:
\begin{eqnarray*}
     minimize & \MObj_{\coefEps}(\setvar)=\oneobj{k}(\setvar) & \\
    \text{subject to} & \oneobj{i}(\setvar) \leq \coefEps_i & i \in [1, \nbobj], i\neq k \\
    & \setvar \in \setcts &    
\end{eqnarray*}

in which $\nbobj \geq 2$ is the number of objective functions, $\oneobj{i}$ the $i^{th}$ objective function and $\setcts$ represents the set of feasible solutions.
An initial point minimizing $\oneobj{k}(\setvar)$ can be determined by setting $\coefEps_i$ to infinity.
Afterwards, by varying the values of $\coefEps_i$, it is possible to compute the set of non-dominated solutions, (\emph{i.e.}, the Pareto front).

\section{Fairness-enhancing Methods}
\label{sec:related}

While many approaches have been proposed in the literature to enhance the fairness of machine learning methods, they can be categorized in three main families, namely \emph{preprocessing} techniques~\cite{kamiran2012data,zemel2013learning,feldman2015certifying,calmon2017optimized}, \emph{algorithmic modification} techniques~\cite{kamiran2010discrimination,calders2010three,kamishima2012fairness,zafar2017fairness} and \emph{postprocessing} techniques~\cite{hardt2016equality}. 
In a nutshell, preprocessing techniques aim at changing the characteristics of the input data (\emph{e.g.}, by removing existing correlations with the sensitive attribute) so that any classifier trained on this data achieves fairness with respect to its prediction. 
In contrast, algorithmic modification techniques integrate the fairness constraints directly into a learning algorithm to ensure that the outputted model is fair. 
Finally, postprocessing techniques modify the outcome of an already trained model to ensure fairness.

Our work falls within the \emph{algorithmic modification} approach in the sense that we propose a fairness-aware algorithm for learning fair rule lists. 
In particular, we integrate fairness constraints into \corels{} to produce rule lists that satisfy several statistical notions of fairness. 
Related solutions include the seminal work of Calders and Verwer (2010)~\cite{calders2010three}, which consists in training as many classifiers as demographic subgroups defined by $A$, before using at test time the classifier associated to a particular subgroup. 
Hereafter, we focus on the related work with respect to training fair \emph{and} interpretable classifiers, which is the objective of our method.

Kamiran, Calders and Pechenizkiy (2010)~\cite{kamiran2010discrimination} have proposed a learning algorithm incorporating the discrimination and accuracy gains into the splitting criterion of a decision tree classifier. 
In particular, they have devised three strategies (\emph{i.e.}, difference, ratio and sum) combining the accuracy and discrimination gains into a single one used as splitting criterion. 
They have also added a leaf relabeling post-processing technique that changes the label of selected leaves to improve fairness. 
In the same line of work, Raff, Sylvester and Mills (2018)~\cite{raff2018fair} have applied the difference-based strategy  from~\cite{kamiran2010discrimination} on CART decision trees~\cite{breiman2017classification} to create fair decision tree and random forest~\cite{breiman2001random}.
Zhang and Ntoutsi (2019)~\cite{zhang2019faht} proposes \faht{}, which improved the performance of combination-based strategies by introducing the so-called \emph{fair information gain}, which corresponds to the default accuracy gain when there is no unfairness, and to the product of the fairness gain and the accuracy gain otherwise. 
In addition, they use a Hoeffding tree~\cite{DBLP:conf/kdd/DomingosH00} to provide better fairness and accuracy in the online setting.

More recently, Zafar, Valera, Gomez-Rodriguez and Gummadi (2019)~\cite{zafar2019fairness} propose a
constraint-based framework to design fair margin-based classifier (\emph{e.g.}, logistic regression and support vector machines) supporting statistical parity, equal opportunity and equalized odds.
This constrained logistic regression \zafar{}~\cite{zafar2019fairness} offers better unfairness/accuracy compare to existing \emph{fair and interpretable} techniques. 
For instance, on the Adult dataset~\cite{frank2010uci}, \faht{}~\cite{zhang2019faht} achieves a statistical parity of $0.16$ for an accuracy of $0.818$, while \zafar{} achieves a statistical parity of $0.073$ for the same accuracy. 

However, \emph{fair and interpretable} techniques usually achieve lower performance compared to state-of-the-art \emph{fair and non-interpretable} techniques. 
For instance, still on Adult, for an accuracy of $\sim0.80$, \zafar{} achieves a statistical parity of $\sim0.05$ while \laftr{}~\cite{madras2018learning} -- which is to the best of our knowledge the state-of-the-art \emph{non-interpretable} technique for fair classification -- displays a statistical parity of $\sim0.025$. 
In this work, we compare the performance of \fairl{} to both \zafar{}~\cite{zafar2019fairness} and \laftr{}~\cite{madras2018learning}, and show that it outperforms the former while being competitive with the latter.

\section{FairCORELS}
\label{sec:fairrulelists}

Given a dataset $\dataset{}=(\legitimate{}, A, \decision{})$, our objective is to learn a rule list model subject to a fairness constraint. 
To compute such model, we extend \corels{} to accept solutions within the branch-and-bound method only when they satisfy the minimal fairness constraint\footnote{Equivalently, our implementation actually \emph{minimizes the misclassification error} with a \emph{maximum acceptable unfairness}.}.

\fairl{} is depicted in Algorithm~\ref{alg:faircorels}, in which the unfairness level is controlled by the parameter $\epsilon$. 
We use $\unfair{}(\arule, \legitimate{}, A,\decision{})$ to denote an oracle measuring the unfairness (using one of the the metrics detailed in Section~\ref{subsec:fairness}) of a rule list $\arule$ given features vectors $\legitimate{}$, sensitive attributes $A$ and labels $\decision{}$. 
We refer the reader to the original \corels{} papers for a detailed understanding of the  algorithm~\cite{angelino2017learning,angelino2018learning}. 
In a nutshell, similarly to \corels{}, \fairl{} represents the search space of the fair rule lists as a trie. 
Then, while the trie contains unexplored leaves, the next prefix $\arule$ to extend is outputted by the search strategy considered. 
For every antecedent $s$ not present in the prefix $\arule$, if the lower bound of the rule list $\arule' = \arule \cup \{s\}$ is less than the current minimum objective, then $\arule'$ is inserted in both the queue and the trie. 
The current minimum objective is updated whenever it is higher than the objective of $\arule'$ and $\arule'$ satisfies the fairness constraint. 
As mentioned in Section~\ref{subsec:rulelist}, \corels{}'s objective function is the sum of the misclassification error of the associated rule list with a regularization term penalizing its length.

\begin{algorithm}
\caption{\fairl{}}
\label{alg:faircorels}
\begin{algorithmic}[1]
\STATE Inputs: 
Training data $(\legitimate{}, \decision{})$; 
set of antecedents $\setrules{}$; 
sensitive attribute $A$; objective function $\obj{}(\arule, \legitimate{}, \decision{})$; objective lower bound $\lwb{}(\arule, \legitimate{},\decision{})$; unfairness oracle $\unfair{}(\arule, \legitimate{}, A,\decision{})$; 
fairness constraint parameter $\coefEps$; 
initial best known rule list $\arule^0$ with objective value $\oneobj{}^0$ and unfairness $\unfair{}(\arule^0,\legitimate{},A,\decision{}) \leq 1-\coefEps$.
\STATE Output: Rule list $\arule^*$ with minimum objective $\oneobj{}^*$ such that $\unfair{}(\arule^*,\legitimate{},A,\decision{}) \leq 1-\coefEps$.
\STATE $(\arule^c, \oneobj{}^c) \gets  (\arule^0, \oneobj{}^0)$
\STATE $Q \gets queue([()])$ 
\WHILE{$Q$ not empty}
    \STATE $\arule \gets Q.pop()$ 
    \IF{$\lwb{}(\arule, \legitimate{}, \decision{}) < \oneobj{}^c$}            
        \STATE $\oneobj{} \gets \obj{}(\arule, \legitimate{}, \decision{})$
        \STATE $u \gets \unfair{}(\arule, \legitimate{}, A, \decision{})$ 
        \IF{$\oneobj{} < \oneobj{}^c$ and $u \leq 1-\coefEps$} 
            \STATE $(\arule^c, \oneobj{}^c) \gets (\arule, \oneobj{})$ 
        \ENDIF
        \FOR {$s$ in $\setrules{}$} 
            \IF{$s$ not in  $\arule$} 
                    \STATE $Q.push((\arule,s))$
            \ENDIF
        \ENDFOR
    \ENDIF
\ENDWHILE
\STATE $(\arule^*, \oneobj{}^*) \gets  (\arule^c, \oneobj{}^c)$
\end{algorithmic}
\end{algorithm}

Using \fairl{} to implement the $\epsilon-$constraint method is straightforward.
Given a set of constraints $E_\epsilon$, for each constraint value $\epsilon \in E_\epsilon$, we compute the rule list satisfying such constraint using \fairl{}. Finally, we return the set of non-dominated solutions.

As our work is an extension of \corels{}, one can exploit its strategies and lower bounds to control the tree search expansion. 
In this work, we have considered three search strategies of \corels{}: the breadth-first search (\texttt{BFS original}) and the best-first with respectively the lower bound (\texttt{Lower bound}) and curiosity (\texttt{Curious}) priorities. 
In addition, we implement a breadth-first search (\texttt{BFS obj.-aware}), which evaluates first the prefixes with higher objective-function value among those of a given length. 
Note that during our preliminary experiments, we also tried other search strategies such as depth-first search as well as best-first search guided by the objective function. 
However, none of these strategies has lead to better solutions while they were having either a higher memory or computational footprints.

\section{Experiments}
\label{sec:experiments}

The aim of experimental analysis is mainly to address three questions: 
(Q1) \emph{How does \fairl{} perform on the six supported statistical fairness notions?} (Q2) \emph{How does \fairl{} compare to state-of-the-art techniques for fair classification?} (Q3) \emph{What does a fair rule list look like?}

\subsection{Experimental Setup}

We focus on two classification problems: (1) predicting which subjects in the \emph{Adult} dataset~\cite{frank2010uci} earn more than $50,000\$$ per year and (2) predicting which subjects in the \emph{COMPAS} dataset~\cite{angwin2016machine} will re-offend within two years.
In all our experiments, \fairl{} searches for rule list formed by single- and two-clause antecedents. 
Furthermore, for all the methods evaluated (including \fairl{}), we assume that the information about the sensitive attribute is not used for making the prediction.

More precisely, the Adult dataset contains information about more than $45,000$ individuals from the $1994$ U.S. census, with the sensitive attribute being the gender (Female/Male).
The pre-processed dataset includes $M=30,300$ records, for which our rule mining procedure yields $N=144$ single- and two-clause antecedents. 
The COMPAS dataset gathers $6,167$ records from criminal offenders in Florida during $2013$ and $2014$, for which we use the race (African-American/Caucasian) as the sensitive attribute. 
When pre-processed, the dataset includes $M=5,200$ records and our rule mining procedure yields $N=166$ single- and two-clause antecedents.

\paragraph{Training Procedure.} For both datasets, we set the maximum size of the trie to $n=4.10^6$. 
This means that during the branch-and-bound, the execution is stopped if the number of nodes in the trie exceeds $n$. For the regularization parameter, we used $\lambda=10^{-3}$. 
To compute the input data used for \fairl{}, we first binarize categorical features and discretize real-valued features. 
For the COMPAS dataset, we used the same discretized dataset used for \corels{}~\cite{angelino2017learning}. 
For the Adult dataset, we used one third of the dataset to learn the splits with the \emph{Minimum Description Length Principle}~\cite{Fayyad1993MultiIntervalDO} and applied the learned splits to the remaining two thirds before using it as discretized dataset. 

The mining procedure takes as input the set $R$ formed by the binarized categorical features and the discretized real-valued features and return $R \cup \bar{R} \cup T$, with $T$ denoting two-clause antecedents that has a minimal support $\sigma$. 
More precisely, $T = \{r_i \wedge r_j | (r_i,r_j)\in R \times R, r_i \neq r_j, supp(r_i \wedge r_j) \geq \sigma  \}$, in which $\sigma=0.1$ (respectively $\sigma=0.01$) for the Adult dataset (respectively the COMPAS dataset). 

Our experiments were conducted on an Intel Xeon Processor E3-1271 v3 (3.60 GHz) with 32GB of RAM. \fairl{} is implemented in C++ and based on the original source code of CORELS\footnote{\url{https://github.com/nlarusstone/corels}}. 
We also build an open source Python package performing the Python binding for CORELS\footnote{\url{https://github.com/fingoldin/pycorels}}.

\subsection{Fairness Metrics Supported by \fairl{}}
\label{sec:results_metrics}

In this section, we evaluate the error/unfairness tradeoffs found by \fairl{} on each of the six statistical notions of fairness supported. 
For each metric, the performance of \fairl{} is evaluated using the four search strategies mentioned in Section~\ref{subsec:rulelist}: \texttt{BFS original}, \texttt{BFS obj.-aware}, \texttt{Curious} and \texttt{Lower Bound}.

\paragraph{Setup.} For each fairness metric and each search strategy, the Pareto fronts are obtained by sweeping over $60$ values of fairness coefficients $\epsilon \in [0, 1]$. 
Then, for each value of $\epsilon$, we run a $5-$fold cross validation and report the average of both the classification error and the unfairness on the test set. Finally, we compute the set of non-dominated points. 

\begin{figure*}[h!]
\centering
\begin{subfigure}[t]{\textwidth}
\centering
\includegraphics[width=\textwidth]{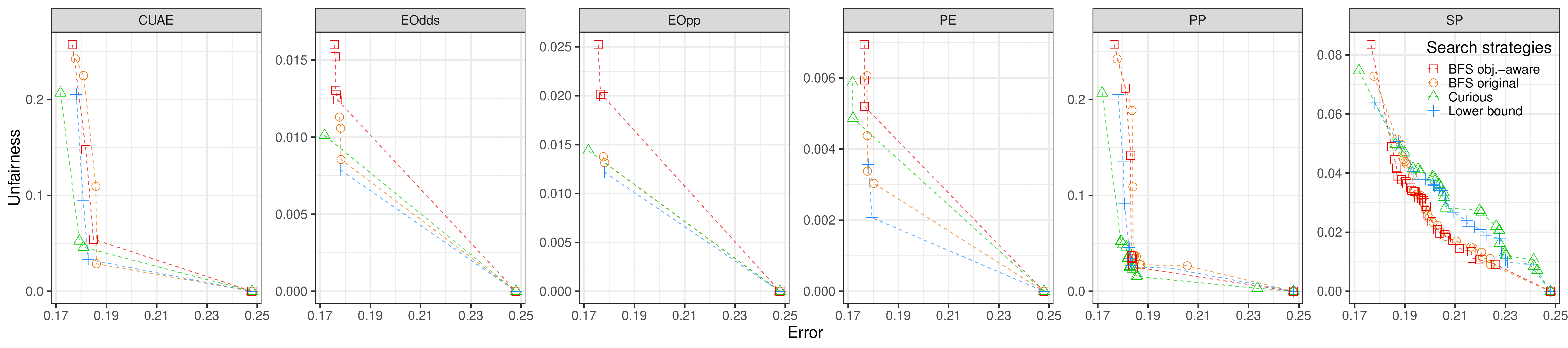}
\caption{Performances of \fairl{} on Adult dataset.}
\label{subfig:metrics_adult}
\end{subfigure}%
\hfill
\begin{subfigure}[t]{\textwidth}
\centering
\includegraphics[width=\textwidth]{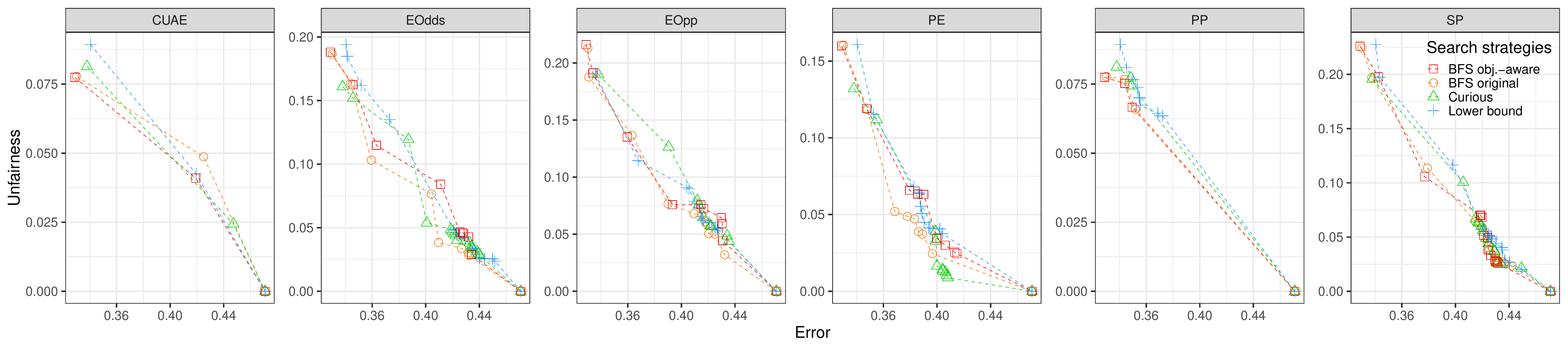}
\caption{Performances of \fairl{} on COMPAS dataset.}
\label{subfig:metrics_compas}
\end{subfigure}%
\caption{Pareto front (unfairness/error tradeoffs) of \fairl{} for statistical parity ($\msp{}$), predictive parity, predictive equality, equal opportunity ($\meopp{}$), equalized odds ($\meodds{}$) and conditional use accuracy equality on Adult and COMPAS datasets. For each fairness metric, we show the performances obtained using four search strategies: BFS objective-aware, BFS, Curious and Lower bound.}
\label{fig:metrics}
\end{figure*}

\paragraph{Results.} Figure~\ref{fig:metrics} shows the error/unfairness tradeoffs of \fairl{} for fair classification on both Adult (Figure~\ref{subfig:metrics_adult}) and COMPAS (Figure~\ref{subfig:metrics_compas}) datasets, evaluated on the six supported metrics, namely statistical parity, predictive parity, predictive equality, equal opportunity, equalized odds and conditional use accuracy equality. 
Overall, our method is able to discover interesting tradeoffs for all the six fairness notions evaluated, thus showing that (1) \fairl{} can effectively explore the error/unfairness tradeoffs and (2) is agnostic to the statistical notion of fairness considered. 
In addition, the analysis of the search strategies provide useful information with respect on how to further improve the performances of \fairl{}. 
For instance, one can use all the four strategies during the cross-validation and select the one yielding the best trade-offs. 

\subsection{Comparison to State-of-the-art Techniques}
\label{sec:results_sota}

We compare the error/unfairness tradeoffs of \fairl{} to those of state-of-the-art non-interpretable ( \laftr{}~\cite{madras2018learning}) and interpretable  (\zafar{}~\cite{zafar2019fairness}) techniques for fair classification on both Adult and COMPAS, using statistical parity ($\unfair{SP}$), equal opportunity ($\unfair{EOpp}$) and equalized odds ($\unfair{EOdds}$) as fairness constraint.
We chose to use these three statistical notions of fairness because they are implemented in both \zafar{}~\cite{zafar2019fairness} and \laftr{}~\cite{madras2018learning}.

\paragraph{Setup.} For the experiments on \laftr{}, we use the code provided by the authors~\cite{madras2018learning}\footnote{\url{https://github.com/VectorInstitute/laftr}}.
For both datasets, we used the same setting as the authors, which is a single-hidden layer neural network with $8$ hidden units for each of the encoder, classifier and adversary.
For the experiments on \zafar{}, we use the code provided by the authors~\cite{zafar2019fairness}\footnote{\url{https://github.com/mbilalzafar/fair-classification}}. 
The fairness constraints are solved using \texttt{DCCP}\footnote{\url{https://github.com/cvxgrp/dccp}} as optimizer. 
We set the optimizer parameters to $(\tau=5.0,\mu=1.2)$ (respectively $(\tau=20.0,\mu=1.2)$) for the Adult (respectively COMPAS) dataset.

For \zafar{} and \fairl{}, the Pareto fronts are obtained by first sweeping over $60$ values of fairness coefficients $\epsilon \in [0, 1]$. 
Then, for each value of $\epsilon$, we run a $5-$fold cross validation and report the average of both the classification error and the unfairness on the test set. 
Finally, we compute the set of non-dominated points. 
In addition, for \fairl{}, for each fairness metric, we use the results of all four search strategies to compute the Pareto front. 

For \laftr{}, we follow a similar validation procedure as the authors~\cite{madras2018learning}. 
This procedure consists in first learning a fair representation of the data, by sweeping across a range of $60$ fairness coefficients $\gamma \in [0.1, 4]$\footnote{In their original work, the authors have tested $10$ coefficients, which provide only a few points on the Pareto front. In contrast, we use $60$ points to be consistent with \fairl{} and \zafar{}.}. 
More precisely, for each value of $\gamma$, an encoder is trained for $1,000$ epochs, with model checkpoints made every $50$ epochs. 
Afterwards, for each checkpoint, $7$ classifiers are trained on the corresponding fair representation using $7$ different random seeds. 
Then, the epoch $e$ with the lowest median error on the validation set is selected. 
Finally, $7$ additional classifiers are trained on an unseen test, using the selected best epoch $e$ and the $7$ random seeds. 
The median of both the error and the unfairness across the $7$ classifiers are used to compute the Pareto front.

\begin{figure*}[h!]
\centering
\begin{subfigure}[t]{0.33\textwidth}
\centering
\includegraphics[width=\textwidth]{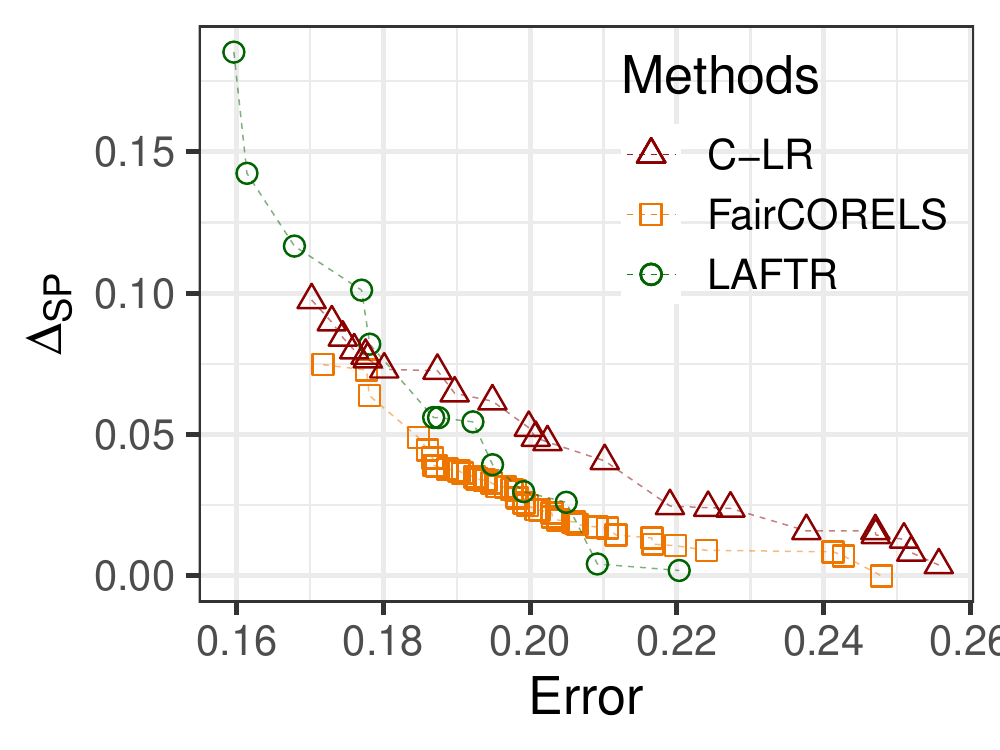}
\caption{Trade-off between error and $\msp{}$}
\label{subfig:adult_sp}
\end{subfigure}%
\hfill
\begin{subfigure}[t]{0.33\textwidth}
\centering
\includegraphics[width=\textwidth]{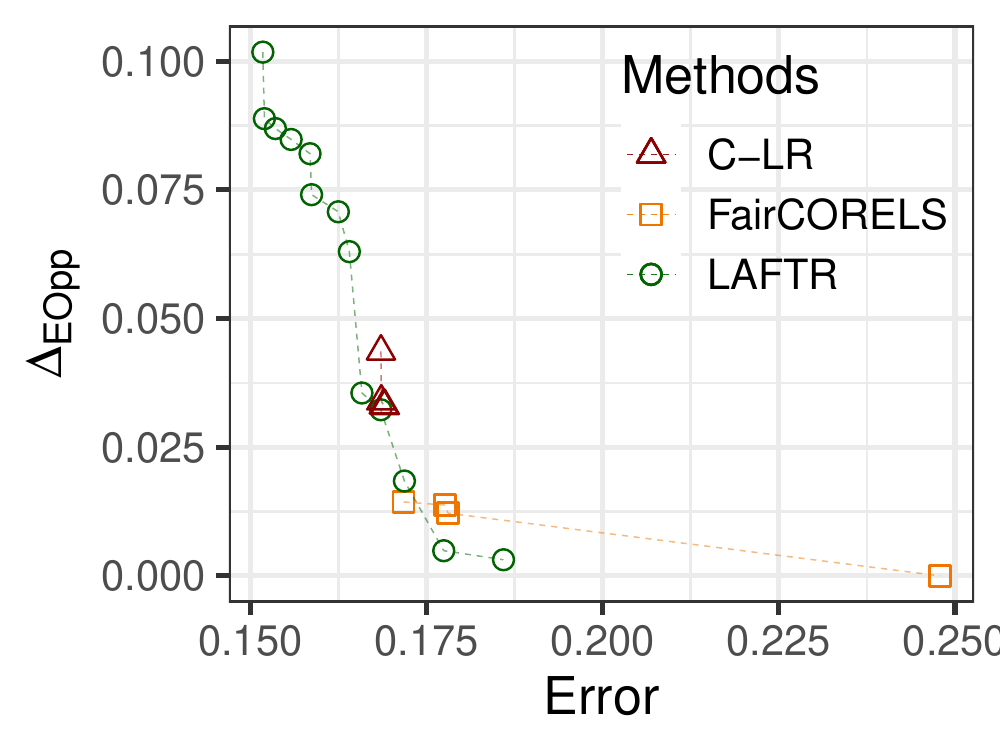}
\caption{Tradeoff between error and $\meopp{}$}
\label{subfig:adult_eopp}
\end{subfigure}%
\hfill
\begin{subfigure}[t]{0.33\textwidth}
\centering
\includegraphics[width=\textwidth]{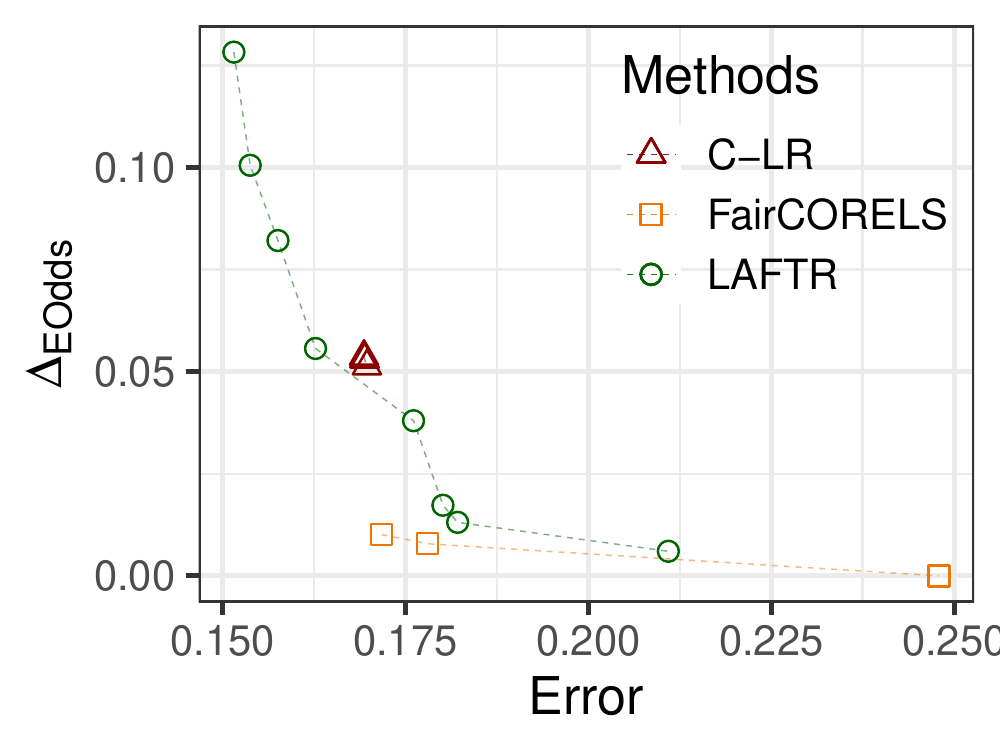}
\caption{Tradeoff between error and $\meodds$}
\label{subfig:adult_odds}
\end{subfigure}%
    \caption{Pareto front (unfairness/ classification error tradeoffs) for statistical parity ($\msp{}$), equal opportunity ($\meopp{}$) and equalized odds ($\meodds{}$) on Adult. Bottom-left (low unfairness, low error) is preferable.}
\label{fig:pareto_adult}
\end{figure*}

\begin{figure*}[h!]
\begin{subfigure}[t]{.33\textwidth}
        \centering
        \includegraphics[width=\linewidth]{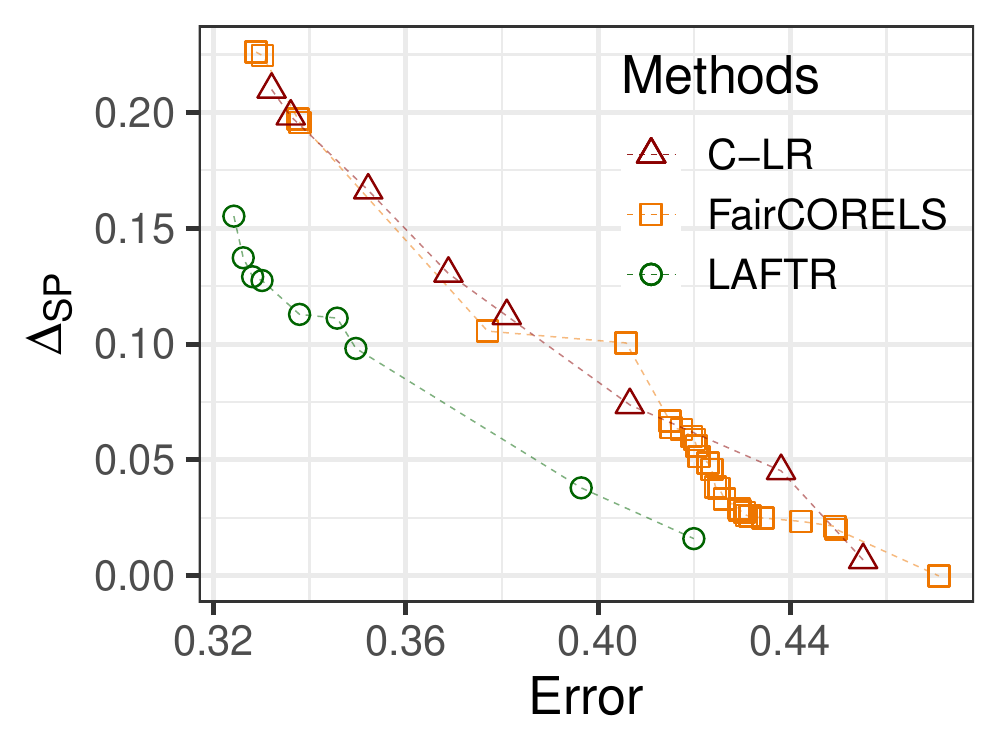}
        \caption{Tradeoff between error and $\msp{}$}
        \label{subfig:compas_sp}
\end{subfigure}\hfill
\begin{subfigure}[t]{.33\textwidth}
        \centering
        \includegraphics[width=\linewidth]{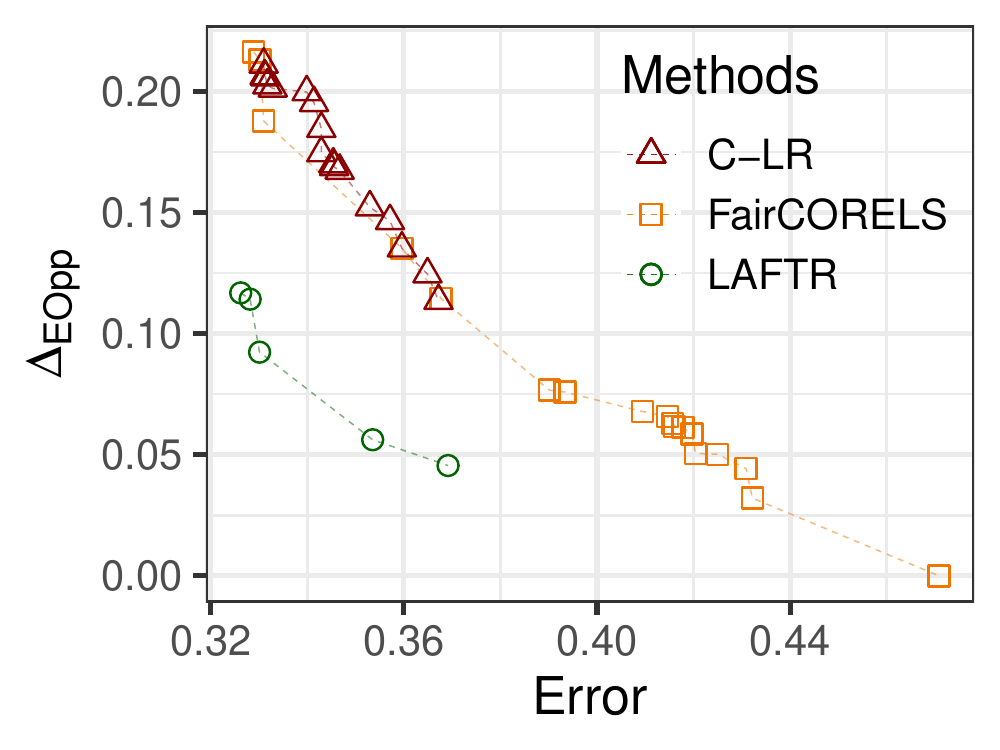}
        \caption{Tradeoff between error and $\meopp{}$}
        \label{subfig:compas_eopp}
\end{subfigure}\hfill
\begin{subfigure}[t]{.33\textwidth}
        \centering
        \includegraphics[width=\linewidth]{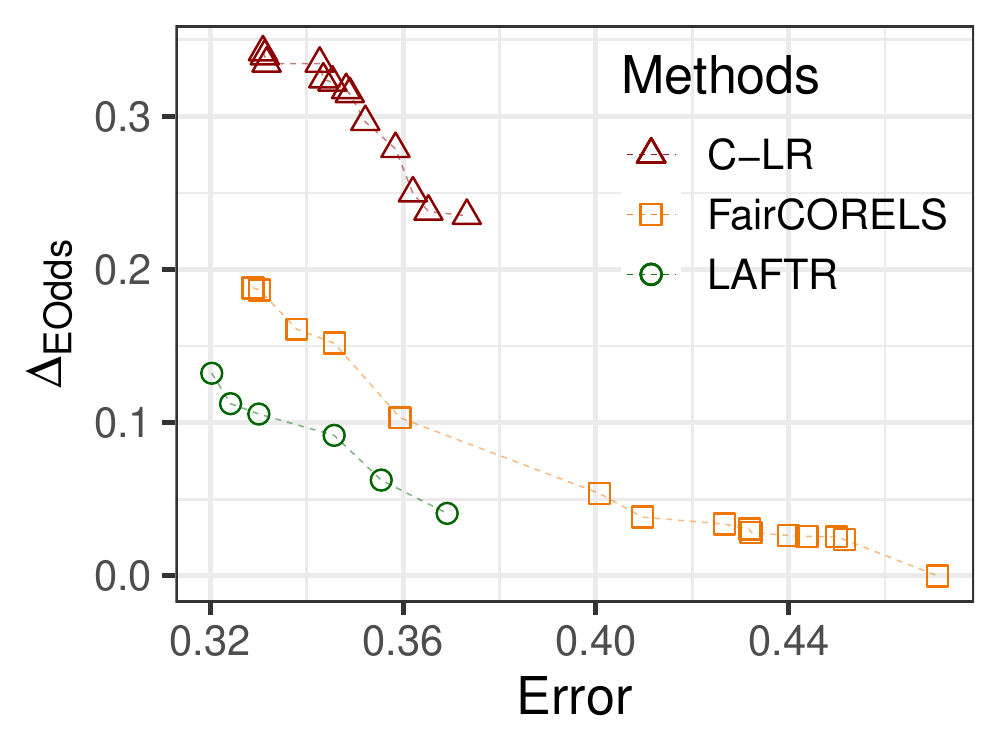}
        \caption{Tradeoff between error and $\meodds$}
        \label{subfig:compas_odds}
\end{subfigure}%
\caption{Pareto front (unfairness/classification error tradeoffs) for statistical parity ($\msp{}$), equal opportunity ($\meopp{}$) and equalized odds ($\meodds{}$) on COMPAS. 
Bottom-left (low unfairness, low error) is preferable.}
\label{fig:pareto_compas}
\end{figure*}

\paragraph{Results.} Figure~\ref{fig:pareto_adult} and~\ref{fig:pareto_compas} show the results of fair classification with \zafar{}~\cite{zafar2019fairness}, \laftr{}~\cite{madras2018learning} and \fairl{}, on Adult and COMPAS. 

Figure~\ref{fig:pareto_adult} shows the comparisons for the Adult dataset. 
Each sub-figure plots the error-unfairness tradeoffs of \fairl{}, \zafar{}, and \laftr{}, evaluated on statistical parity (Fig.~\ref{subfig:adult_sp}), equal opportunity (Fig.~\ref{subfig:adult_eopp}) and equalized odds (Fig.~\ref{subfig:adult_odds}). 
Figure~\ref{subfig:adult_sp} shows that, for statistical parity, in lower unfairness regime (unfairness $\leq 0.05$), -- which is the desirable regime of fair classification -- \fairl{} offers better performances than \zafar{} while being competitive with \laftr{}. 
In fact, there is only one point (error $=0.21$) at which \fairl{} is not better or similar to \laftr{}. 
For this particular point, we observe a statistical parity of $0.004$ (respectively $0.017$) for \laftr{} (respectively \fairl{}). 
For both equal opportunity and equalized odds, the plots in Figure~\ref{subfig:adult_eopp} and Figure~\ref{subfig:adult_odds} show that \fairl{} offers the best performances in lower unfairness regime, followed by \laftr{}{} and then by \zafar{}.

Figure~\ref{fig:pareto_compas} describes the results obtained for the COMPAS dataset. 
For statistical parity, Figure~\ref{subfig:compas_sp} shows that \laftr{} offers the best performances, followed by \fairl{} and then by \zafar{}. 
Notice that while \fairl{} performances are very close to those of \laftr{} in lower regime of unfairness (unfairness $\leq 0.05$). 
For both equal opportunity and equalized odds, the plots in Figure~\ref{subfig:compas_eopp} and Figure~\ref{subfig:compas_odds} show that only \fairl{} was able to find solutions in lower regime of unfairness.

\subsection{Examples of Fair Rule Lists Found by \fairl{}}

\paragraph{Adult dataset.}
Rule lists~\ref{rl:rl_adult_uncons}, \ref{rl:rl_adult_sp} and \ref{rl:rl_adult_eopp} illustrate some results returned by \fairl{} on Adult dataset, namely an unconstrained rule list, a rule list with statistical parity constraint and a rule list with equal opportunity constraint. 
Examples for all implemented fairness notions are shown in Appendix~\ref{app:adult}.

\begin{minipage}{\linewidth}
\begin{lstlisting}[language=Haskell,numbers=none, caption={Example of an unconstrained rule list found by \fairl{} on Adult dataset, with $\acc = 0.829$, $\unfair{SP} = 0.091$ and $\unfair{EOpp}=0.013$.}
,label=rl:rl_adult_uncons]
if [capitalGain:>=5119] then [high]
else if [maritalStatus:single] then [low]
else if [workclass:private &&  education:bachelors] then [high]
else if [education:masters_doctorate] then [high]
else if [capitalLoss:1882-1978] then [high]
else [low]
\end{lstlisting}
\end{minipage}

\begin{minipage}{\linewidth}
\begin{lstlisting}[language=Haskell,numbers=none, caption={Example of a \emph{SP-constrained} rule list found by \fairl{} on Adult dataset, with $\acc = 0.801$ and $\unfair{SP} = 0.019$.}
,label=rl:rl_adult_sp]
 if [capitalGain:>=5119] then [high]
else if [occupation:not_professional] then [low]
else if [hoursPerWeek:>50 && workclass:private] then [low]
else if [age:36-61] then [high]
else [low]
\end{lstlisting}
\end{minipage}

\begin{minipage}{\linewidth}
\begin{lstlisting}[language=Haskell,numbers=none, caption={Example of an \emph{EOpp-constrained} rule list found by \fairl{} on Adult dataset, with $\acc = 0.823$ and $\unfair{EOpp} = 0.008$.}
,label=rl:rl_adult_eopp]
 if [capitalGain:>=5119] then [high]
else if [maritalStatus:single] then [low]
else if [capitalLoss:1882-1978] then [high]
else if [workclass:private && education:bachelors] then [high]
else if [education:masters_doctorate] then [high]
else [low]
\end{lstlisting}
\end{minipage}

\paragraph{COMPAS dataset.} Rule lists~\ref{rl:rl_compas_uncons}, \ref{rl:rl_compas_odds} and \ref{rl:rl_compas_cuae} show some rule lists found by \fairl{} on COMPAS, namely an unconstrained rule list, a rule list with equalized odds constraint and a rule list with conditional use accuracy equality. 
Examples for all fairness notions are shown in Appendix~\ref{app:compas}.

\begin{minipage}{\linewidth}
\begin{lstlisting}[language=Haskell,numbers=none, caption={Example of an unconstrained rule list found by \fairl{} on COMPAS dataset, with $\acc = 0.681$, $\unfair{EOdds} = 0.217$ and $\unfair{CUAE}=0.046$}
,label=rl:rl_compas_uncons]
if [priors:>3] then [recidivism]
else if [age:21-22 && gender:Male] then [recidivism]
else if [age:18-20] then [recidivism]
else if [age:23-25 && priors:2-3] then [recidivism]
else [no recidivism]
\end{lstlisting}
\end{minipage}
\begin{minipage}{\linewidth}
\begin{lstlisting}[language=Haskell,numbers=none, caption={Example of an \emph{EOdds-constrained} rule list found by \fairl{} on COMPAS dataset, with $\acc = 0.591$ and $\unfair{EOdds} = 0.048$.}
,label=rl:rl_compas_odds]
if [juvenile-misdemeanors:0 && priors:1] then [no recidivism]
else if [juvenile-crimes:0 && age:26-45] then [no recidivism]
else if [juvenile-crimes:0 && priors:0] then [no recidivism]
else [recidivism]
\end{lstlisting}
\end{minipage}
\begin{minipage}{\linewidth}
\begin{lstlisting}[language=Haskell,numbers=none, caption={Example of a \emph{CUAE-constrained} rule list found by \fairl{} on COMPAS dataset, with $\acc = 0.645$ and $\unfair{CUAE} = 0.009$.}
,label=rl:rl_compas_cuae]
if [age:18-20 && gender:Male] then [recidivism]
else if [priors:>3] then [recidivism]
else if [age:21-22 && gender:Male] then [recidivism]
else if [age:23-25 && priors:2-3] then [recidivism]
else [no recidivism]
\end{lstlisting}
\end{minipage}

\section{Conclusion}
\label{sec:conclusion}

In this paper, we presented \fairl{}, a fairness-aware algorithm to learn fair and interpretable models by design. 
We formulated the problem of learning fair rule lists as a bi-objective formulation of the problem of learning rule list, in which we jointly minimize the unfairness as well as the classification error.  
\fairl{} is embedded in a bi-objective optimization method to compute the set of non-dominated solutions. 
One strength of our approach is that it is agnostic to the statistical notion of fairness considered. 
Our experiments show that this technique can identify better fairness/accuracy trade-offs than previous existings works.

As for any branch-and-bound algorithm, \fairl{} relies on lower and upper bounds to efficiently prune the search space. Designing efficient unfairness bounds is an interesting and important future research goal in order to improve the performances of \fairl{}. We also plan to develop custom branching strategies that could lead to better error/unfairness tradeoffs.

\bibliographystyle{icml2020}
\bibliography{papers}

\clearpage
\appendix
\section{Examples of rule lists found by \fairl{} on Adult dataset}
\label{app:adult}
\begin{minipage}{\linewidth}
\begin{lstlisting}[language=Haskell,numbers=none, caption={Example of an unconstrained rule list found by \fairl{} on Adult dataset, with $\acc = 0.829$, $\unfair{SP} = 0.091$, $\unfair{PP} = 0.265$, $\unfair{PE} = 0.007$, $\unfair{EOpp}=0.013$, $\unfair{EOdds}=0.010$ and $\unfair{CUAE}=0.265$ .}
,label=rl:rl_adult_uncons_2]
if [capitalGain:>=5119] then [high]
else if [maritalStatus:single] then [low]
else if [workclass:private &&  education:bachelors] then [high]
else if [education:masters_doctorate] then [high]
else if [capitalLoss:1882-1978] then [high]
else [low]
\end{lstlisting}
\end{minipage}

\begin{minipage}{\linewidth}
\begin{lstlisting}[language=Haskell,numbers=none, caption={Example of a \emph{SP-constrained} rule list found by \fairl{} on Adult dataset, with $\acc = 0.801$ and $\unfair{SP} = 0.019$.}
,label=rl:rl_adult_sp_2]
 if [capitalGain:>=5119] then [high]
else if [occupation:not_professional] then [low]
else if [hoursPerWeek:>50 && workclass:private] then [low]
else if [age:36-61] then [high]
else [low]
\end{lstlisting}
\end{minipage}

\begin{minipage}{\linewidth}
\begin{lstlisting}[language=Haskell,numbers=none, caption={Example of a \emph{PP-constrained} rule list found by \fairl{} on Adult dataset, with $\acc = 0.819$ and $\unfair{PP} = 0.036$.}
,label=rl:rl_adult_pp]
 if [age:<21] then [low]
else if [capitalGain:>7073] then [high]
else if [capitalLoss:2385-2581] then [high]
else if [not_capitalLoss:1882-1978] then [low]
else if [maritalStatus:married] then [high]
else [low]
\end{lstlisting}
\end{minipage}
\begin{minipage}{\linewidth}
\begin{lstlisting}[language=Haskell,numbers=none, caption={Example of a \emph{PE-constrained} rule list found by \fairl{} on Adult dataset, with $\acc = 0.828$ and $\unfair{PE} = 0.001$.}
,label=rl:rl_adult_pe]
 if [capitalGain:>5119] then [high]
else if [maritalStatus:single] then [low]
else if [capitalLoss:1882-1978] then [high]
else if [education:masters_doctorate] then [high]
else if [education:prof_school] then [high]
else [low]
\end{lstlisting}
\end{minipage}

\begin{minipage}{\linewidth}
\begin{lstlisting}[language=Haskell,numbers=none, caption={Example of an \emph{EOpp-constrained} rule list found by \fairl{} on Adult dataset, with $\acc = 0.823$ and $\unfair{EOpp} = 0.008$.}
,label=rl:rl_adult_eopp_2]
 if [capitalGain:>=5119] then [high]
else if [maritalStatus:single] then [low]
else if [capitalLoss:1882-1978] then [high]
else if [workclass:private && education:bachelors] then [high]
else if [education:masters_doctorate] then [high]
else [low]
\end{lstlisting}
\end{minipage}

\begin{minipage}{\linewidth}
\begin{lstlisting}[language=Haskell,numbers=none, caption={Example of an \emph{EOdds-constrained} rule list found by \fairl{} on Adult dataset, with $\acc = 0.823$ and $\unfair{EOdds} = 0.006$.}
,label=rl:rl_adult_eodds]
 if [capitalGain:>5119] then [high]
else if [maritalStatus:single] then [low]
else if [capitalLoss:1882-1978] then [high]
else if [hoursPerWeek:<34] then [low]
else if [education:masters_doctorate] then [high]
else [low]
\end{lstlisting}
\end{minipage}

\begin{minipage}{\linewidth}
\begin{lstlisting}[language=Haskell,numbers=none, caption={Example of a \emph{CUAE-constrained} rule list found by \fairl{} on Adult dataset, with $\acc = 0.823$ and $\unfair{CUAE} = 0.056$.}
,label=rl:rl_adult_cuae]
 if [capitalGain:5317-7073] then [low]
else if [capitalGain:>5119] then [high]
else if [capitalLoss:1882-1978] then [high]
else if [capitalLoss:1821-1862] then [high]
else if [capitalLoss:2385-2581] then [high]
else [low]
\end{lstlisting}
\end{minipage}
\section{Examples of rule lists found by \fairl{} on COMPAS dataset}
\label{app:compas}

\begin{minipage}{\linewidth}
\begin{lstlisting}[language=Haskell,numbers=none, caption={Example of an unconstrained rule list found by \fairl{} on COMPAS dataset, with $\acc = 0.681$, $\unfair{SP}=0.272$, $\unfair{PP}=0.046$, $\unfair{PE}=0.222$, $\unfair{EOpp}=0.212$, $\unfair{EOdds} = 0.217$ and $\unfair{CUAE}=0.046$}
,label=rl:rl_compas_uncons_2]
if [priors:>3] then [recidivism]
else if [age:21-22 && gender:Male] then [recidivism]
else if [age:18-20] then [recidivism]
else if [age:23-25 && priors:2-3] then [recidivism]
else [no recidivism]
\end{lstlisting}
\end{minipage}

\begin{minipage}{\linewidth}
\begin{lstlisting}[language=Haskell,numbers=none, caption={Example of a \emph{SP-constrained} rule list found by \fairl{} on COMPAS dataset, with $\acc = 0.581$ and $\unfair{SP} = 0.040$.}
,label=rl:rl_compas_sp]
 if [juvenile-crimes:0 && age:26-45] then [no recidivism]
else if [age:23-25 && priors:1] then [no recidivism]
else if [juvenile-crimes:0 && priors:0] then [no recidivism]
else [recidivism]
\end{lstlisting}
\end{minipage}

\begin{minipage}{\linewidth}
\begin{lstlisting}[language=Haskell,numbers=none, caption={Example of a \emph{PP-constrained} rule list found by \fairl{} on COMPAS dataset, with $\acc = 0.642$ and $\unfair{PP} = 0.005$.}
,label=rl:rl_compas_pp]
if [age:26-45 && priors:>3] then [recidivism]
else if [juvenile-crimes:0 && age:18-20] then [recidivism]
else if [juvenile-misdemeanors:0 && juvenile-felonies:0] then [no recidivism]
else if [age:26-45] then [no recidivism]
else [recidivism]
\end{lstlisting}
\end{minipage}

\begin{minipage}{\linewidth}
\begin{lstlisting}[language=Haskell,numbers=none, caption={Example of a \emph{PE-constrained} rule list found by \fairl{} on COMPAS dataset, with $\acc = 0.597$ and $\unfair{PP} = 0.056$.}
,label=rl:rl_compas_pe]
if [charge_degree:Misdemeanor && age:26-45] then [no recidivism]
else if [priors:0 && charge_degree:Felony] then [no recidivism]
else if [juvenile-misdemeanors:0 && priors:1] then [no recidivism]
else [recidivism]
\end{lstlisting}
\end{minipage}

\begin{minipage}{\linewidth}
\begin{lstlisting}[language=Haskell,numbers=none, caption={Example of an \emph{EOpp-constrained} rule list found by \fairl{} on COMPAS dataset, with $\acc = 0.585$ and $\unfair{EOpp} = 0.072$.}
,label=rl:rl_compas_eopp]
 if [juvenile-crimes:>0 && gender:Male] then [recidivism]
else if [priors:>3 && gender:Female] then [recidivism]
else if [priors:2-3 && charge_degree:Felony] then [recidivism]
else [no recidivism]
\end{lstlisting}
\end{minipage}

\begin{minipage}{\linewidth}
\begin{lstlisting}[language=Haskell,numbers=none, caption={Example of an \emph{EOdds-constrained} rule list found by \fairl{} on COMPAS dataset, with $\acc = 0.591$ and $\unfair{EOdds} = 0.048$.}
,label=rl:rl_compas_odds_2]
if [juvenile-misdemeanors:0 && priors:1] then [no recidivism]
else if [juvenile-crimes:0 && age:26-45] then [no recidivism]
else if [juvenile-crimes:0 && priors:0] then [no recidivism]
else [recidivism]
\end{lstlisting}
\end{minipage}

\begin{minipage}{\linewidth}
\begin{lstlisting}[language=Haskell,numbers=none, caption={Example of a \emph{CUAE-constrained} rule list found by \fairl{} on COMPAS dataset, with $\acc = 0.645$ and $\unfair{CUAE} = 0.009$.}
,label=rl:rl_compas_cuae_2]
if [age:18-20 && gender:Male] then [recidivism]
else if [priors:>3] then [recidivism]
else if [age:21-22 && gender:Male] then [recidivism]
else if [age:23-25 && priors:2-3] then [recidivism]
else [no recidivism]
\end{lstlisting}
\end{minipage}

\end{document}